\documentclass[runningheads]{llncs}

\usepackage[review,year=2024,ID=2721]{eccv}

\usepackage{eccvabbrv}

\usepackage{graphicx}
\usepackage{booktabs}
\usepackage{amsmath,amssymb,amsfonts}
\usepackage{graphicx}
\usepackage{textcomp}
\usepackage{algorithm2e}
\usepackage{framed,multirow}
\usepackage{subcaption}
\usepackage{tabularx}
\usepackage{latexsym}
\usepackage{array}
\usepackage[accsupp]{axessibility}  

\usepackage[pagebackref,breaklinks,colorlinks,citecolor=eccvblue]{hyperref}

\usepackage{orcidlink}

\begin{document}

\title{Dataset Distillation for Histopathology Image Classification} 

\titlerunning{Abbreviated paper title}

\author{Cong Cong\inst{1}\and
Shiyu Xuan\inst{2,3}\and
Sidong Liu\inst{1}\and 
Maurice Pagnucco\inst{4}\and
Shiliang Zhang\inst{3}\and
Yang Song\inst{4}}

\authorrunning{C.Cong et al.}

\institute{Australian Institute of Health Innovation (AIHI),
Macquarie University, Macquarie Park, NSW, 2109, Australia \and
School of Computer Science and Engineering, Nanjing
University of Science and Technology, Nanjing 210094, China \and
National Key Laboratory for Multimedia Information Processing, School of Computer Science, Peking University, Beijing, 100871, China \and
School of Computer Science and Engineering, University of New South Wales, Australia}

\maketitle

\begin{abstract}
Deep neural networks (DNNs) have exhibited remarkable success in the field of histopathology image analysis.
On the other hand, the contemporary trend of employing large models and extensive datasets has underscored the significance of \emph{dataset distillation}, which involves compressing large-scale datasets into a condensed set of synthetic samples, offering distinct advantages in improving training efficiency and streamlining downstream applications.
In this work, we introduce a novel dataset distillation algorithm tailored for histopathology image datasets (Histo-DD), which integrates stain normalisation and model augmentation into the distillation progress. Such integration can substantially enhance the compatibility with histopathology images that are often characterised by high colour heterogeneity.
We conduct a comprehensive evaluation of the effectiveness of the proposed algorithm and the generated histopathology samples in both patch-level and slide-level classification tasks. 
The experimental results, carried out on three publicly available WSI datasets, including Camelyon16, TCGA-IDH, and UniToPath, demonstrate that the proposed Histo-DD can generate more informative synthetic patches than previous coreset selection and patch sampling methods.
Moreover, the synthetic samples can preserve discriminative information, substantially reduce training efforts, and exhibit architecture-agnostic properties. These advantages indicate that synthetic samples can serve as an alternative to large-scale datasets.
  \keywords{Histopathology \and Image classification \and Dataset distillation}
\end{abstract}

\section{Introduction}
\label{sec:intro}
Histopathology image analysis plays a pivotal role in delivering critical insights for cancer diagnosis and prognosis \cite{khened2021generalized}. The advent of advanced image scanning technology, particularly Whole-Slide Imaging (WSI), which digitises entire histopathology slides, has spurred considerable interest among researchers to develop computer-aided systems in this domain. 
Notably, deep learning-based approaches have emerged as a focal point of investigation across various applications, encompassing histopathology image preprocessing (\emph{e.g.}, techniques for staining normalisation \cite{de2021residual,cong2022colour}) and subsequent analyses, including histopathology image classification \cite{qu2022dgmil,yang2022remix,zhang2022gigapixel}, segmentation \cite{graham2023one,guo2023sac}, and survival prediction \cite{fan2022cancer}.

Given the immense size of WSIs, it is impractical to input an entire WSI into a model for training. Therefore, most previous approaches partition WSIs into smaller patches and subsequently consolidate the results from these patches to derive slide-level conclusions. Thus, a cluster of studies has emerged with the aim of enhancing patch-level analysis \cite{li2018cancer,shen2020deformable,cong2022imbalanced}. Nonetheless, to ensure the accuracy of patch-level analysis, precise annotations at the patch level are typically required. 
However, patch-level annotations are unavailable in many scenarios, and only slide-level labels can be accessed. Consequently, an alternative line of research has emerged; treating the analysis of WSIs as a weakly-supervised learning task. Many of these methodologies rely on Multiple Instance Learning (MIL) frameworks \cite{chen2022scaling,qu2022dgmil,zhao2022setmil,yang2022remix,shao2021transmil}. 
Though effective, these methods necessitate training on vast numbers of patches, imposing significant computational and storage demands. One natural question to ask is: \textit{Do we really need such huge amounts of patches to obtain an effective model for WSI analysis?} In fact, many MIL-based models already adopt some patch selection criteria during feature aggregation. These methods \cite{chikontwe2020multiple,yao2020whole,yang2022remix} select a small group of key patches based on their importance to the task or feature similarities. 
Although these strategies are effective, the criteria of key patch selection are often heuristic. 
In contrast, our objective is to distill the vast patch dataset using a deep learning approach, so that the original dataset can be represented by a very small set of synthetic samples while minimising information loss. 

To obtain a highly condensed dataset, coreset selection methods have been introduced \cite{phillips2017coresets,mirzasoleiman2020coresets}. These methods aim to curate a small set of samples using heuristics such as representation compactness \cite{rebuffi2017icarl} and forgetfulness \cite{toneva2018empirical}. While these methods offer computational efficiency, they encounter challenges, such as the absence of a guarantee that the selected samples are representative or optimal for downstream tasks \cite{zhao2020dataset}. 
More recently, \cite{wang2018dataset} proposed dataset distillation to synthesise informative samples such that the model trained on them can perform similarly to that trained on the full training set. 
However, despite the success of dataset distillation methods in demonstrating performance guarantees using synthetic samples, their applicability to histopathology datasets has remained largely unexplored. This can be attributed to the unique challenges posed by histopathology images. Firstly, histopathology slides are incredibly large, often reaching gigapixel sizes. Secondly, these images can vary significantly in terms of tissue structures, staining, and cellular morphology. 

To this end, we introduce a dataset distillation algorithm for histopathology images (Histo-DD).
Rather than directly generating synthetic slides, Histo-DD generates samples from cropped patches. The synthetic patches, derived from the same slides, can be aggregated to create synthetic slides for slide-level analysis.
The core idea of Histo-DD is that the model trained on the synthesised patches should provide similar performances to the one trained on the full dataset. 
To do this, we explicitly minimise the differences in layer-wise gradients between the model trained on synthetic samples and its counterpart trained on the complete dataset.
To enhance the resilience of synthetic samples across various models and optimise training efficiency, we adopt model augmentation \cite{zhang2023accelerating} to train synthetic samples. 
However, the inherent heterogeneity in stain colours can pose a challenge to effectively training synthetic samples. Drawing inspiration from the stain normalisation methods \cite{macenko2009method,cong2022colour}, we propose integrating these techniques to mitigate this issue.
One way to achieve this objective is to normalise the dataset as a pre-processing step, however, this strategy only brings marginal performance gains (shown in Section \ref{sec:ab}). 
In contrast, we resort to Differentiable Siamese Augmentation \cite{zhao2021dataset} and propose to integrate a differentiable stain normalisation as outlined in \cite{macenko2009method}, along with other standard data augmentation techniques into the training procedure. Our results demonstrate that this integration substantially enhances the quality of the distilled samples and once these samples are learned, they can be used with stain normalisation algorithm to address colour variation issues in histopathology images.

We illustrate that the distilled samples obtained by Histo-DD can be seamlessly used for downstream patch and slide-level tasks, delivering comparable performance to models trained on the full dataset while largely reducing the training effort and only using less than 1\% of the entire patch population. 
Furthermore, we explore the utility of our distilled samples in a real-life application: cross-centre transfer learning. 
Results on the TCGA-IDH \cite{liu2020isocitrate}, Camelyon16 \cite{bejnordi2017diagnostic} and UniToPath \cite{2021UniToPatho} datasets demonstrate that the synthetic patches retain most discriminative information contained in the original large-scale patch datasets, making them a viable substitute for the latter.
To our knowledge, this study represents the first attempt to design a dataset distillation method for histopathology images, opening up new avenues for research and applications in this domain.
Specifically, our contributions are summarised as follows:
\begin{itemize}
\item We present the first study in dataset distillation for histopathology images, where we leverage a deep learning-based approach to generate a highly condensed set of synthetic patches. 
Compared to heuristic-based patch sampling or coreset selection methods, such learned synthetic samples offer a more encompassing representation of the rich information contained in the original dataset since they are directly optimised for the downstream tasks.
Due to their small scale, these synthetic samples can then be efficiently utilised to train various deep-learning models for downstream analysis. 

\item We propose to integrate differentiable stain normalisation into the learning process of synthetic patches. This enhancement improves the compatibility of dataset distillation algorithms specifically tailored for histopathology images.
\item We show a broad usage of the synthetic samples including patch-level classification, slide-level classification, and cross-centre in-domain transfer learning. Using synthetic samples on these tasks demonstrates a significant improvement in efficiency and obtains promising results.
\end{itemize}

\section{Related Work}
\subsection{Coreset Selection}
Coreset selection methods aim to find the most informative subset of the entire dataset, in which the model trained on it has close performance with the one trained on the full dataset.
Selection of coreset instances can be conducted according to various criteria. For example, geometry-based approaches select key samples based on their distance to the coreset centre in the feature space \cite{welling2009herding,farahani2009facility}.
Moreover, error-based methods are based on the assumption that samples contributing more to the model's loss are crucial for training. Accordingly, samples that result in higher errors or gradients are selected for training \cite{toneva2018empirical,paul2021deep}.
Furthermore, recent studies \cite{ducoffe2018adversarial,margatina2021active} tend to select samples that are closer to the decision boundary, as they are considered more challenging to classify.
It is worth noting that most of these methods rely on heuristic selection criteria for incrementally choosing important data points.
However, these heuristics cannot guarantee the optimality of the selected subset for training machine learning models \cite{wang2018dataset}.

\subsection{Dataset Distillation}
Dataset distillation aims to reduce the size of a large-scale dataset to a small set of synthetic samples, such that the models trained on these synthetic samples can perform similarly to the ones trained on the original large dataset \cite{wang2018dataset}. Unlike coreset selection, dataset distillation modifies the selected samples, thus removing the restrictions imposed by uneditable elements and preserving a greater amount of information \cite{lei2023comprehensive}.
Based on different learning approaches, the current dataset distillation methods can be roughly divided into two categories. 
The first category approaches dataset distillation as a meta-learning task. In this context, the distilled data are formulated as learnable parameters that can be refined through a nested optimisation loop. The nested loop encompasses an extensive inner loop that unrolls an optimisation graph, which is notably computationally demanding \cite{wang2018dataset}. 
Deng \emph{et al.} \cite{deng2022remember} improved this unrolling process by adding a momentum term for extending the length of unrolled trajectories. Moreover, Nguyen \emph{et al.} \cite{nguyen2020dataset} approximate the inner-loop model optimisation by posing the classification task as a ridge regression problem. The synthetic set serves as the support set, while the original set acts as the target set. This method is further improved by Loo \emph{et al.} \cite{loo2022efficient} which adopts a more ﬂexible conjugate kernel for approximating the output of a neural network.

Rather than optimising directly with respect to model performance, the second group of works uses \emph{surrogate objective approaches} that focus on optimising an alternative objective. For example, Zhao \emph{et al.} \cite{zhao2020dataset} propose to synthesise samples by minimising the layer-wise gradient differences. They also propose to optimise synthetic samples such that their data distribution is similar to that of real samples \cite{zhao2023dataset}. 
Moreover, the CAFE framework \cite{wang2022cafe} employs a layer-wise feature alignment technique to enhance the learning of synthetic samples.
Beyond these two primary categories, there are several other notable efforts in dataset distillation. Zhao \emph{et al.} \cite{zhao2021dataset} explore the effectiveness of data augmentation in synthetic sample learning, and Zhang \emph{et al.} \cite{zhang2023accelerating} propose Acc-DD which adopts model augmentation to largely improve the training efficiency of dataset distillation algorithms. Recently, Liu \emph{et al.} \cite{liu2023dream} proposed the DREAM framework which adopts clustering algorithms to select only representative samples for matching during optimisation, and Sajedi \emph{et al.} \cite{sajedi2023datadam} learn synthetic images by aligning the spatial attention maps of real and synthetic data.

Recently, dataset distillation algorithms have also been studies in the medical image domain. For example, Tian \emph{et al.} \cite{tian2023communication} propose a generalisable dataset distillation algorithm to reduce communication costs in federated learning for skin lesion analysis. Moreover, Li \emph{et al.} \cite{li2020soft,li2022compressed} adopt a gradient matching algorithm to generate condensed sets of X-ray or gastric images. Their research demonstrates that these condensed images exhibit significantly enhanced efficiency and improved security when sharing data. However, the application of dataset distillation algorithms has not been explored in histopathology images.

\section{Methods}
We present a dataset distillation algorithm (Histo-DD) designed to generate synthetic samples from a WSI dataset. 
Given a set of $N$ WSI slides each containing $T_{n}$ sampled patches with $n$ denoting the slide's index, depending on the given task, our algorithm can be applied directly on these $\sum_{1}^{N}T_{n}$ patches to obtain a highly condensed size of $m$ synthetic patches per class for a patch-level classification task, with $m\ll \sum_{1}^{N}T_{n}$.
Additionally, we can apply our algorithm to each of these $N$ slides in which we distil from $T_{n}$ patches to just $m$ synthetic patches per class (\emph{i.e.}, $N\times m$ synthetic patches in total) for a slide-level classification task.
The overall algorithm is described in Algorithm \ref{alg:1} in which $\mathcal{P}$ denotes the set of synthetic patches and $\mathcal{D}$ denotes the original patches.  
By optimising the patch classifier, we establish a link between the optimisation objectives for both the original and distilled datasets, facilitating knowledge distillation from $\mathcal{D}$ to $\mathcal{P}$.
To mitigate the challenges posed by high colour heterogeneity in histopathology datasets, we introduce stain normalisation as a data augmentation technique to further enhance the quality of the learned synthetic patches. 
\SetKwComment{Comment}{/* }{ */}
\RestyleAlgo{ruled}
\begin{algorithm}
\caption{Histo-DD algorithm for learning synthetic histopathology patches.}\label{alg:two}\label{alg:1}
\KwData{Training set $\mathcal{D}$, random initialised synthetic samples $\mathcal{P}$, a set of pre-trained models $\mathcal{S}$, differentiable augmentation $Aug(\cdot)$, \# of distillation steps $K$, \# of steps for model updates $U$, and the number of classes $C$.}
\For{$k=0,\cdots,K-1$}{
    Randomly select $\mathcal{M}_{\theta}$ from $\mathcal{S}$\;
    Perform model augmentation on $\mathcal{M}_{\theta}$\;
    \For{$u=0,\cdots,U-1$}{
        \For{$c=0,\cdots C$}{
            Sample a batch $\mathcal{D}^c$, $\mathcal{P}^c$ from $\mathcal{D}$ and $\mathcal{P}$, respectively\;
            $\mathcal{D}_A^c \gets Aug(\mathcal{D}^c)$\; 
            $\mathcal{P}_A^c \gets Aug(\mathcal{P}^c)$\; 
            Calculate loss $L_{\mathcal{D}_A^c} \gets \mathcal{L}(\mathcal{D}_A^c,\mathcal{M}_{\theta})$\;
            Calculate loss $L_{\mathcal{P}_A^c} \gets \mathcal{L}(\mathcal{P}_A^c,\mathcal{M}_{\theta})$\;
            Update $\mathcal{P}_A^c \gets \mathop{\min}_{\mathcal{P}_A^c}\sum_{l=0}^{L}Dis(\nabla_{k} L_{\mathcal{D}_A^c},\nabla_{k} L_{\mathcal{P}_A^c})$\;
        }
        Update $\mathcal{M}_{\theta} \gets \mathcal{L}(\mathcal{D},\mathcal{M}_{\theta})$\;
    }
}
\end{algorithm}

\subsection{Synthetic Patch Learning}
Recent studies \cite{zhao2021dataset,zhao2020dataset} show that gradient matching has attained state-of-the-art performance in the task of dataset distillation. 
Consequently, in this study, we employ gradient matching as our optimisation algorithm which adopts a nested loop optimisation to learn $\mathcal{P}$ from $\mathcal{D}$.
Specifically, the optimisation includes an outer loop (OL), which runs for $K$ steps, for synthetic sample optimisation, and an inner loop (IL), which runs for $U$ steps, for updating the classification model. 

In each OL, a classification model $\mathcal{M}_{\theta}$ (\emph{e.g.,} ResNet, ConvNet, \emph{etc.}) is chosen. To ensure the learned synthetic samples can generalise well to different $\mathcal{M}_{\theta}$, previous studies require a significant number of iterations in OL. 
This can pose a notable challenge for histopathology images due to the abundance of patches derived from slides, resulting in a substantial increase in training iterations. Thus, we employ model augmentation \cite{zhang2023accelerating} in OL which ensures the generalisation capacity of the synthetic samples while largely reducing training efforts.

In each loop of IL, we first sample a batch of real samples ($\mathcal{D}^{c}$) and synthetic samples ($\mathcal{P}^{c}$) from each class $c$. 
To improve the generalisation performance of $\mathcal{D}$, we apply Differentiable Siamese Augmentation (DSA) \cite{zhao2021dataset} during training. Meanwhile, to effectively tackle the high colour heterogeneity issue of histopathology images, we further integrate stain normalisation into the DSA process. 
Specifically, given a differentiable augmentation operation $Aug(\cdot)$, 
we obtain $\mathcal{D}_A^{c}$ and $\mathcal{P}_A^{c}$ via $Aug(\mathcal{D}^{c})$ and $Aug(\mathcal{P}^{c})$. 
Then $\mathcal{P}_A^{c}$ are optimised using the following equation:
\begin{equation}
\mathop{\min}_{\mathcal{P}_A^{c}}\sum_{c=0}^{C}Dis(\nabla_{k}\mathcal{L}(\mathcal{D}_A^{c},\mathcal{M}_{\theta}),\nabla_{k}\mathcal{L}(\mathcal{P}_A^{c},\mathcal{M}_{\theta}))
    \label{equ:gm}
\end{equation}
where $C$ is the total number of layers of a model, $Dis(\cdot,\cdot)$ is a distance function and $\mathcal{L}(\cdot,\cdot)$ denotes the loss function used for train the $\mathcal{M}_{\theta}$. In our implementation, we use $L$1 loss for calculating $Dis(\cdot,\cdot)$ and cross-entropy for $\mathcal{L}(\cdot,\cdot)$.

\paragraph{Model augmentation} Given the typically vast number of patches in histopathology datasets, naively adopting gradient match algorithms is time-consuming since the size of OL is usually large to ensure the generalization ability of synthetic samples. Drawing inspiration from \cite{zhang2023accelerating}, we employ model augmentation to enhance the generalization capacity of synthetic samples while significantly reducing the need for a high number of OLs.
Specifically, we first pre-train a set ($\mathcal{S}$) of networks with different hyperparameters.
In detail, we train a patch-level classifier 5 times with different weight initialisations and then collect these five trained models to form $\mathcal{S}$. 
During the outer loop, we randomly select one pre-trained model ($\mathcal{M}_{\theta}$) from $\mathcal{S}$.
Then we apply model augmentation via perturbing the model parameter $\theta$ with a small value $d$ sampled from a Gaussian distribution $\mathcal{N}(0,1)$.
We normalise $d$ to make it compatible with different filters at different layers:
\begin{equation}
    d_{l,j}\leftarrow\frac{d_{l,j}}{\left \| d_{l,j} \right \|_F}\left \| w_{l,j} \right \|_F
    \label{equ:d}
\end{equation}
where $l,j$ denotes the $j$-th filter in $l$-th layer of $d$, $w_{l,j}$ denotes the corresponding parameters and  $\left \| \cdot \right \|$ denotes the Frobenius norm. Then, we can apply $d$ to $\theta$ via:
\begin{equation}
    \theta\leftarrow \theta+\alpha\times d
    \label{equ:theta}
\end{equation}
and we use $\alpha$ to control the degree of perturbation. Applying model augmentation reduces the required number of outer loops which improves efficiency.

\paragraph{Data augmentation with stain normalisation} To further improve the generalisation performance of synthetic samples and alleviate the high colour heterogeneity issue of histopathology images, we integrate DSA with stain normalisation. 
By solving Eq. \ref{equ:gm}, we derive the following equation:
\begin{equation}
    \frac{\partial Dis(\cdot)}{\partial \mathcal{P}_A^{c}}=\frac{\partial Dis(\cdot)}{\partial \nabla_{k}\mathcal{L}(\cdot)}\cdot \frac{\partial \nabla_{k}\mathcal{L}(\cdot)}{\partial Aug(\cdot)}\cdot \frac{\partial Aug(\cdot)}{\partial \mathcal{P}_A^{c}}
    \label{equ:diff}
\end{equation}
Eq. \ref{equ:diff} shows that $Aug(\cdot)$ needs to be differentiable with regards to $\mathcal{P}_A^{c}$. 
Following DSA, standard data augmentation can be reformulated as differentiable functions for signal back-propagation to $\mathcal{P}_A^{c}$.
However, it is implicit to reformulate existing stain normalisation approaches into differentiable functions compatible with DSA in a memory-efficient manner. 
Although trained deep learning-based stain normalisation methods \cite{shaban2019staingan,salehi2020pix2pix,cong2021texture,cong2022colour} can be used as black-box models for normalisation, they are sub-optimal as they necessitate additional model training and extra memory to use them during training.
In contrast, conventional stain normalisation methods \cite{reinhard2001color,macenko2009method,vahadane2016structure}, which pose stain normalisation as a stain separation task, can be used without training.
Thus, to optimise for both computational efficiency and differentiability, we opt for a conventional approach \cite{macenko2009method} and implement a differentiable version to fit into the training process. 
In particular, we chose a template image whose pixel mean and standard deviation are closest to the overall mean and standard deviation of the dataset and then computed its stain vector and colour concentration offline. 
During training, stain normalisation augmentation simply involves matrix multiplication to normalise the image with the template colour appearances using the obtained stain vector and concentration, ensuring overall differentiability.

\section{Experiments}
\subsection{Dataset Details}
We trained and evaluated our proposed dataset distillation method (Histo-DD) on three medical datasets,
including TCGA-IDH \cite{liu2020isocitrate}, Camelyon16 \cite{bejnordi2017diagnostic}, and UniToPath \cite{2021UniToPatho}. 
Further details regarding the quantity of train/test images can be found in the supplementary material.

\noindent\textbf{TCGA-IDH} \cite{liu2020isocitrate} is used for predicting the isocitrate dehydrogenase (IDH) gene mutation status as it is an important diagnostic, prognostic, and therapeutic biomarker in glioma \cite{parsons2008integrated}. 
The dataset comprises two classes: Mutant (MU) and Wild Type (WT). 
We split the slides into train/test partitions following \cite{liu2020isocitrate}.
The training set comprises 1,191 slides, with 477 being MU and 714 being WT. 
Additionally, the testing dataset includes 59 MU slides and 90 WT slides.
Within each slide, patches were cropped at a size of 1024$\times$1024, captured at a 10$\times$ magnification level. Only patches with over 50\% tissue content were considered for both training and evaluation. The ground truth labels for these patches were derived from the slide-level labels.

\noindent\textbf{Camelyon16} \cite{bejnordi2017diagnostic} is used for classifying breast histopathology as normal or tumour.
It contains 399 slides of which 270 slides (159 normal vs 111 tumour) are used for training and 129 slides (80 normal vs 49 tumour) are used for testing. 
Specifically, we first apply the Otsu algorithm \cite{otsu1979threshold} to filter out the background regions of each slide and then densely extract patches
of size 256$\times$256 pixels at 5$\times$ magnification level.
Since the detailed pixel annotations are provided, we have labelled patches extracted from normal regions as `normal' and those from tumour regions as `tumour'.

\noindent\textbf{UniToPath} \cite{2021UniToPatho} is used for colorectal polyps type classification (PT) and adenoma dysplasia grading (DG). 
It contains 9,536 patches extracted from 292 WSIs at 20$\times$ magnification.
For PT, each slide is annotated by expert UniTo pathologists, and is classified into six classes:
normal tissue (NORM), hyperplastic polyp (HP), tubular adenoma high-grade dysplasia (TA.HG), tubular adenoma low-grade dysplasia (TA.LG), 
tubulo-villous adenoma, high-grade dysplasia (TVA.HG), and tubulo-villous adenoma, low-grade dysplasia (TVA.LG).
We follow the official splits which use 208 slides (13 NORM, 31 HP, 16 TA.HG, 100 TA.LG, 13 TVA.HG and 30 TVA.LG) for training and 88 slides (8 NORM, 10 HP, 9 TA.HG, 46 TA.LG, 7 TVA.HG and 8 TVA.LG) for testing.
Moreover, for DG, which is a binary classification task to predict the grade (LG or HG) of TA and TVA adenomas. We have 160 training slides (130 LG and 30 HG) and 70 testing slides (54 LG and 16 HG).
\subsection{Experimental Setup}
To evaluate the effectiveness of our dataset distillation algorithm, we compared it with coreset selection methods on the patch classification task and we provide ablation studies about the usefulness of stain normalisation in learning synthetic samples from a histopathology dataset.
Moreover, to more thoroughly assess the efficacy of Histo-DD, we have carried out comprehensive experiments focusing on the application of synthetic patches. These experiments encompassed a broad range of applications including
cross-architecture evaluation, patch-level classification, slide-level classification, and in-domain cross-centre transfer learning.
We conduct our experiments using a two-stage approach.
In the first stage, we focus on learning synthetic samples from the training datasets. Following previous works \cite{zhao2020dataset,zhao2023dataset,lu2021data,zhang2023accelerating}, we employed ResNetAP-10 as our choice for dataset distillation. 
ResNetAP-10 is a modified version of ResNet-10, where the strided convolution is replaced with average pooling for downsampling. 
This modification enables more detailed gradients for synthetic images, leading to an enhancement in the quality of synthetic samples.
In the second stage, we use the learned synthetic samples to train a classification model and test it on the original test set. 
We apply data augmentation for training in both stages.
By evaluating the performance of the real test set, we aim to explore the effectiveness of synthetic patches in capturing the most representative characteristics of the original dataset.

To be consistent with the standard measurements in prior studies \cite{zhao2020dataset,zhao2023dataset}, for each application, we repeated the experiment 5 times with different seeds and reported the mean and standard deviation of classification accuracy and the average running time for each epoch. Moreover, for the majority of our experiments, we investigated three distinct settings involving 10, 20, or 50 synthetic patches per class. This selection of quantities is based on their proven effectiveness from prior studies \cite{zhao2020dataset,zhao2023dataset,zhao2021dataset,wang2022cafe,sajedi2023datadam} in illustrating learning efficiency, which is the primary objective of dataset distillation. 

We randomly select patches from the original dataset as the initial states of synthetic samples. 
For 10, 20, and 50 synthetic patches per class, we use 200, 500, and 1000 distillation steps ($K$) correspondingly and use a consistent value of 100 for model update steps ($U$).
For TCGA-IDH, we learn the synthetic samples using a learning rate of $5e^{-3}$, whereas for other datasets, we use a learning rate of $0.5$. The discussion of these hyper-parameter selections is provided in Section \ref{sec:ab} and the supplementary.
For patch-level classifiers, we trained the models for 50 epochs, and for slide-level MIL models, we trained them for 60 epochs.
We conducted all experiments using a workstation with a single NVidia Tesla V100 GPU and the codes were developed in Python (v3.8) using PyTorch (v1.10).
\begin{table*}[th!]
\centering
\scriptsize
\begin{tabular}{c|cc|ccc|c|c}
\hline
\multicolumn{1}{l|}{} & \multicolumn{1}{l}{Img/Cls} & \multicolumn{1}{l|}{Ratio\%} & \multicolumn{1}{l}{Random} & \multicolumn{1}{l}{Herding} & \multicolumn{1}{l|}{Forgetting} & \multicolumn{1}{l|}{Histo-DD} & \multicolumn{1}{l}{Full} \\ \hline
\multirow{3}{*}{TCGA-IDH}  & 10                           &  $6e^{-2}$                         &   {$51.9_{\pm0.02}$}                         &  {$53.0_{\pm0.7}$}                           &  {$53.0_{\pm3.7}$}           &     {$\mathbf{69.7_{\pm0.2}}$}                       & \multirow{3}{*}{$80.0_{\pm0.7}$}    \\
                      & 20                          &  0.11                          &   {$55.4_{\pm0.01}$}                         &   {$53.4_{\pm0.1}$}                          &   {$54.3_{\pm2.0}$}                          &     {$\mathbf{70.3_{\pm0.1}}$}                      &                          \\
                      & 50                          &  0.28                          &  {$57.1_{\pm0.01}$}                          &    {$54.6_{\pm0.04}$}                         &    {$57.0_{\pm2.7}$}                         &     {$\mathbf{71.8_{\pm0.4}}$}                      &                          \\ \hline
\multirow{3}{*}{Camelyon16}  & 10   &  $9e^{-4}$&   {$68.2_{\pm0.11}$}                         &       {$72.7_{\pm0.40}$}                    &             {$64.2_{\pm0.3}$}                    &             {$\mathbf{87.8_{\pm0.1}}$}               & \multirow{3}{*}{$95.8_{\pm0.05}$}    \\
                      & 20                          &$2e^{-3}$&   {$73.1_{\pm0.04}$}                         &         {$78.4_{\pm1.8}$}                 &       {$69.1_{\pm0.04}$}                 &         {$\mathbf{88.8_{\pm0.2}}$}                  &                          \\
                      & 50                          & $5e^{-3}$&  {$77.6_{\pm0.6}$}                         &    {$81.4_{\pm0.04}$}                        &        {$73.8_{\pm0.1}$}                    &             {$\mathbf{89.6_{\pm1.1}}$}              &                          \\ \hline
\multirow{3}{*}{\begin{tabular}[c]{@{}c@{}}UniToPath\\ (DG)\end{tabular}}  & 10                           &  0.18                          &   {$53.3_{\pm0.03}$}                         &  {$56.4_{\pm0.01}$}                          &  {$58.0_{\pm0.01}$}                                &     {$\mathbf{69.5_{\pm0.8}}$}                       & \multirow{3}{*}{$81.4_{\pm2.3}$}    \\
                      & 20                          &  0.36                          &   {$56.8_{\pm0.01}$}                         &   {$57.2_{\pm0.01}$}                         &   {$59.4_{\pm0.01}$}                         &       {$\mathbf{71.2_{\pm0.8}}$}                   &                          \\
                      & 50                          &  0.89                          &  {$58.7_{\pm0.02}$}                          &    {$60.8_{\pm0.02}$}                        &    {$61.4_{\pm0.02}$}                         &          {$\mathbf{72.1_{\pm1.1}}$}                 &                          \\ \hline
\multirow{3}{*}{\begin{tabular}[c]{@{}c@{}}UniToPath\\ (PT)\end{tabular}}   & 10                           &  0.10                          &   {$29.5_{\pm0.9}$}                         &  {$36.6_{\pm0.1}$}                          &  {$16.7_{\pm0.1}$}                                &        {$\mathbf{43.6_{\pm0.1}}$}                   & \multirow{3}{*}{$49.6_{\pm0.3}$}    \\
                      & 20                          &  0.20                          &   {$32.1_{\pm2.9}$}                         &   {$37.2_{\pm1.2}$}                        &   {$21.5_{\pm0.4}$}                         &             {$\mathbf{44.2_{\pm1.2}}$}             &                          \\
                      & 50                          &  0.52                          &  {$36.8_{\pm0.3}$}                          &    {$39.2_{\pm0.2}$}                       &    {$25.4_{\pm0.3}$}                         &        {$\mathbf{45.2_{\pm0.2}}$}                 &                          \\ \hline

\end{tabular}
\caption{Patch-level classification results on IDH, UniToPath and Camlyon16. We compared our dataset distillation method with different core-set selection methods. The reported results are based on a ResNet18 model.}
\label{table:core}
\end{table*}
\section{Results}
\subsection{Comparison to coreset selection methods}
\label{sec:dd_alg}
We compare our dataset distillation algorithm with coreset selection methods, including Random selection, Herding \cite{welling2009herding} and Forgetting \cite{toneva2018empirical}, in patch classification tasks on three datasets and the results are shown in Table \ref{table:core}.

\begin{figure}[!t]
\centering
\includegraphics[scale=.35]{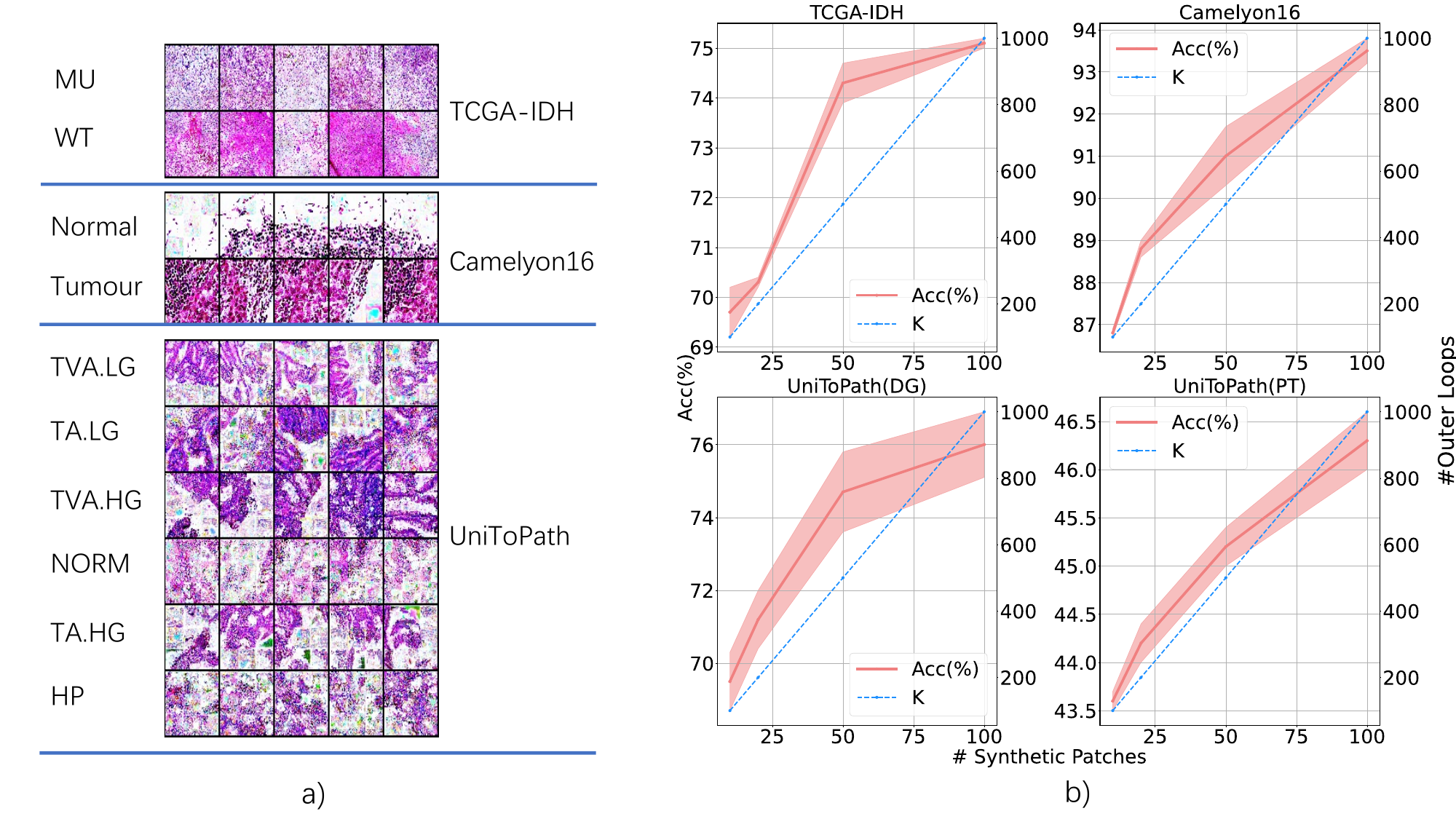}
\caption{a) Visualisation of condensed image patches, and b) Testing accuracy on different datasets using different numbers of synthetic patches per class and their corresponding number of outer loops on convergence.}
\label{fig:vis_number}
\end{figure}

Specifically, for Random selection, we randomly selected samples from the training set as the coreset.
In contrast, the Herding selection method selects coreset samples based on their proximity to the centre of the cluster, while the Forgetting method identifies samples that are more prone to being misclassified by the model. For Herding and Forgetting methods, we use ResNet18 for feature selection.
The results in Table \ref{table:core} show that our dataset distillation method outperforms all compared coreset selection methods by a large margin ($>$10\%). 
Notably, we observed that the smallest performance gap between Histo-DD and the upper bound was on the Camelyon16 dataset. We attribute this to the considerably larger number of patches extracted from Camelyon16, which likely provides a better representation of the real data distribution, thereby yielding more informative learned synthetic samples.
Furthermore, we visualise the synthetic patches for each dataset in Figure \ref{fig:vis_number} (a). 
These synthetic patches exhibit certain histopathological features while also concealing or modifying specific patient-related information. 
\begin{table*}[]
\scriptsize
\centering
\begin{tabular}{cclcccll}
\hline
                                                                        & Img/Cls & None & Stain Norm \cite{macenko2009method} & Rotate & Flip & Scale & All \\ \hline
TCGA-IDH                                                                     & 10      & {$66.6_{\pm0.2}$}&{$67.9_{\pm0.1}$}&{$67.6_{\pm0.1}$}&{$67.0_{\pm0.1}$}&{$67.6_{\pm0.2}$}& {$\mathbf{69.7_{\pm0.2}}$}\\\hline
Camelyon16                                                              & 10      & {$82.9_{\pm0.3}$}     & {$85.5_{\pm0.2}$} &  {$83.1_{\pm0.1}$}      &  {$83.3_{\pm0.3}$}    & $84.2_{\pm0.5}$& {$\mathbf{86.8_{\pm0.1}}$}\\\hline
\begin{tabular}[c]{@{}c@{}}UniToPath\\ (DG)\end{tabular} & 10      & {$64.5_{\pm0.5}$} &{$67.4_{\pm0.4}$}& {$66.1_{\pm0.3}$}&{$66.1_{\pm0.2}$}& {$65.6_{\pm0.4}$}& {$\mathbf{69.5_{\pm0.8}}$}\\\hline
\begin{tabular}[c]{@{}c@{}}UniToPath\\ (PT)\end{tabular}        & 10      & {$39.4_{\pm0.3}$}&{$41.7_{\pm0.1}$}&{$40.3_{\pm0.9}$}& {$41.1_{\pm0.7}$}& {$41.1_{\pm0.8}$}& {$\mathbf{43.6_{\pm0.1}}$}\\ \hline
\end{tabular}
\caption{Patch-level performance with different augmentation strategies.}
\label{table:aug}
\end{table*}
\subsection{Ablation study}
\label{sec:ab}
\noindent\textbf{Effectiveness of stain normalisation.} We demonstrated the effectiveness of various data augmentation techniques in enhancing the learning of synthetic patches, as illustrated in Table \ref{table:aug}. 
Here, we included the results of not applying data augmentations (None), using data augmentations including stain normalisation, rotation, flipping, and scaling, and combining them together (All). Note that we applied the same set of data augmentation to both real samples $\mathcal{D}$ and synthetic samples $\mathcal{P}$ during training.
The results indicate that the incorporation of data augmentation consistently improves the quality of synthetic samples. 
Notably, the application of stain normalisation yields the most substantial enhancement in performance ($\sim 3\%$), underscoring its valuable contribution to the learning process. 
Furthermore, the combination of all data augmentation methods yields the best overall performance. 
Moreover, we investigate training synthetic samples on datasets that are already stain-normalised. 
On the TCGA-IDH dataset, we obtain $64.9_{\pm0.2}$/$67.1_{\pm0.3}$ test accuracy with/without applying stain normalisation on synthetic samples during testing (more results can be found in the supplementary). This indicates that applying stain normalisation as a pre-processing step only brings marginal performance gains. Moreover, it indicates that the synthetic samples learned in this manner are not compatible with stain normalisation techniques, making them less advantageous when addressing stain variations in histopathology datasets.

\noindent\textbf{Different number of synthetic samples.}
Here, we investigate the performance of patch classification tasks on three datasets of a ResNet18 model trained using 10/20/50/100 synthetic patches per class.
As depicted in Figure \ref{fig:vis_number} (b), our findings reveal that classification accuracy improves as the size of the synthetic dataset increases. For instance, when utilizing 100 synthetic patches per class, Histo-DD achieves an accuracy of $75.1\%\pm0.1$, $93.5\%\pm0.3$, $76.0\%\pm0.9$, and $46.3\%\pm0.3$ on TCGA-IDH, Camelyon16, UniToPath(DG), and UniToPath(PT), respectively. 
It is noteworthy that while the performance gap narrows with a larger synthetic dataset, achieving convergence requires more iterations in the outer loop, \emph{i.e.,} larger $K$.
To provide context, each outer loop iteration takes approximately 6 minutes with $U=100$.
Therefore, a trade-off exists between performance improvement and training time. 
Longer training time is indeed a common challenge in current dataset distillation algorithms. 
However, once the synthetic samples are obtained, they can be employed efficiently and effectively as replacements for the original datasets.
\subsection{Synthetic patch evaluation}
\begin{table*}[ht!]
\centering
\scriptsize 
\begin{tabular}{ccc|cccccccc}
\hline
&                          &                       & \multicolumn{2}{c}{RN18}    & \multicolumn{2}{c}{DN121}    & \multicolumn{2}{c}{ViT}                     \\ \hline
\multicolumn{1}{c|}{}  & Img/Cls         & Ratio(\%)     & Acc(\%)        & Time(s)      & Acc(\%)        & Time(s)        & Acc(\%)        & Time(s)                \\ \hline
\multicolumn{1}{c|}{\multirow{4}{*}{TCGA-IDH}}& 10 & $6e^{-2}$&{$69.7_{\pm0.5}$}& $0.24_{(\textcolor{green}{\times167})}$&{$64.0_{\pm0.1}$}& $1.43_{(\textcolor{green}{\times76})}$ & {$71.8_{\pm0.2}$}& $1.84_{(\textcolor{green}{\times98})}$  \\
\multicolumn{1}{c|}{}    & 20 & 0.11 & {$70.3_{\pm0.1}$}&$0.39_{(\textcolor{green}{\times103})}$& {$69.4_{\pm0.1}$} &$2.80_{(\textcolor{green}{\times39})}$ & {$72.5_{\pm0.4}$} & $3.63_{(\textcolor{green}{\times50})}$\\
\multicolumn{1}{c|}{}   & 50 & 0.28 & {$71.8_{\pm0.4}$}&$1.02_{(\textcolor{green}{\times39})}$& {$70.2_{\pm0.4}$} & $4.40_{(\textcolor{green}{\times25})}$ & {$73.4_{\pm0.4}$} & $8.02_{(\textcolor{green}{\times23})}$ \\
\multicolumn{1}{c|}{}   & Full & 100 & {$80.0_{\pm0.7}$}&40 & {$76.3_{\pm0.6}$} & 80 &{$78.9_{\pm0.2}$} & 181\\ \hline
\multicolumn{1}{c|}{\multirow{4}{*}{Camelyon16}}   & 10   & $1e^{-3}$& {$87.8_{\pm0.1}$}  &$0.23_{(\textcolor{green}{\times7043})}$&{$93.5_{\pm0.3}$}& $1.44_{(\textcolor{green}{\times3351})}$ &{$93.9_{\pm0.7}$}&$1.84_{(\textcolor{green}{\times6000})}$\\
\multicolumn{1}{c|}{}    & 20 &$2e^{-3}$&  {$88.8_{\pm0.2}$}  &$0.38_{(\textcolor{green}{\times4263})}$&{$94.0_{\pm0.1}$}&$2.76_{(\textcolor{green}{\times1750})}$&{$94.4_{\pm0.3}$}&$3.61_{(\textcolor{green}{\times3060})}$\\
\multicolumn{1}{c|}{}    & 50  & $5e^{-3}$&  {$89.6_{\pm0.7}$}  &$1.04_{(\textcolor{green}{\times1557})}$&{$95.6_{\pm0.2}$}&$4.40_{(\textcolor{green}{\times1097})}$&{$96.3_{\pm0.3}$}&$8.01_{(\textcolor{green}{\times1380})}$\\
\multicolumn{1}{c|}{}   & Full & 100 & {$95.8_{\pm0.1}$} &  1620&{$98.7_{\pm0.1}$}&4826&{$98.6_{\pm0.1}$}&11049\\ \hline
\multicolumn{1}{c|}{\multirow{4}{*}{\begin{tabular}[c]{@{}c@{}}UniToPath\\ (DG)\end{tabular}}} & 10 & 0.18 &{$69.5_{\pm0.8}$} &$0.25_{(\textcolor{green}{\times320})}$&{$70.5_{\pm1.6}$}&$1.42_{(\textcolor{green}{\times81})}$ &  {$67.2_{\pm0.1}$}&$1.82_{(\textcolor{green}{\times66})}$  \\
\multicolumn{1}{c|}{}  & 20  &0.36  & {$71.2_{\pm0.8}$}&$0.38_{(\textcolor{green}{\times211})}$ &{$74.1_{\pm0.4}$}& $2.80_{(\textcolor{green}{\times41})}$&{$68.6_{\pm0.3}$} &$3.63_{(\textcolor{green}{\times33})}$\\
\multicolumn{1}{c|}{} & 50  & 0.89&  {$72.1_{\pm1.1}$} &$1.01_{(\textcolor{green}{\times79})}$&{$77.0_{\pm0.1}$}&$4.41_{(\textcolor{green}{\times26})}$ &{$73.5_{\pm0.4}$} &$8.07_{(\textcolor{green}{\times15})}$ \\
\multicolumn{1}{c|}{} & Full &100& {$81.4_{\pm2.3}$}& 80&{$82.6_{\pm0.5}$}   & 115 &{$75.8_{\pm0.4}$}   & 120 \\ \hline
\multicolumn{1}{c|}{\multirow{4}{*}{\begin{tabular}[c]{@{}c@{}}UniToPath\\ (PT)\end{tabular}}}   & 10   &   0.10 & {$43.6_{\pm0.1}$} &$0.63_{(\textcolor{green}{\times317})}$& {$46.9_{\pm1.3}$} & $2.01_{(\textcolor{green}{\times105})}$  &{$44.1_{\pm0.9}$}&$5.46_{(\textcolor{green}{\times39})}$ \\
\multicolumn{1}{c|}{}  & 20 &   0.20 & {$44.2_{\pm0.2}$} &$1.16_{(\textcolor{green}{\times172})}$ & {$49.4_{\pm0.8}$} & $3.98_{(\textcolor{green}{\times53})}$  &{$45.6_{\pm0.6}$}&  $10.83_{(\textcolor{green}{\times20})}$   \\
\multicolumn{1}{c|}{}  & 50  &  0.52 & {$44.6_{\pm0.2}$} & $2.93_{(\textcolor{green}{\times68})}$&{$50.5_{\pm0.9}$}                    &$9.88_{(\textcolor{green}{\times21})}$& {$47.1_{\pm0.5}$} &$27.20_{(\textcolor{green}{\times8})}$\\
\multicolumn{1}{c|}{}   & Full                     &   100                    &        {$49.6_{\pm0.2}$}               &      200  &   {$51.5_{\pm0.2}$}                      &        212            &  {$48.5_{\pm0.6}$}                     &        213              \\ \hline
\end{tabular}
\caption{Cross-architecture performance (\%) on different datasets with different synthetic images per class. The distilled samples are obtained using ResNetAP-10 and we also measure the averaged per-epoch training time.}
\label{table:cross}
\end{table*}

\noindent\textbf{Cross-architecture evaluation.}
We evaluated the performance of our condensed data on architectures different from the one used for training them and the results are shown in Table \ref{table:cross}.
Here, we selected five commonly used networks for image classification, including ResNet18 (RN18) \cite{he2016deep}, DenseNet121 (DN121) \cite{huang2017densely} and ViT \cite{dosovitskiy2020image}.
Remarkably, the synthetic samples, even when trained on a different model, can be used to train other models to yield performance that is on par with models trained on the complete dataset.
Notably, training using our synthetic samples is significantly more time-efficient. For instance, training a ViT model on the original Camelyon16 dataset for a single epoch requires a substantial 11,049 seconds. In contrast, training the same ViT model on 50 synthetic samples per class takes only 8.01 seconds for one epoch, which is approximately 1,380 times faster.
Moreover, it's worth emphasising that the synthetic samples constitute less than 1\% of the dataset. 
\begin{table*}[ht!]
\centering
\scriptsize
\begin{tabular}{c|cc|cccccc}
\hline
\multicolumn{1}{l|}{}  & \multicolumn{1}{l}{Img/Cls} & Storage(G)&\multicolumn{1}{l}{CLAM}     & Time(s) & \multicolumn{1}{l}{DSMIL}   & Time(s)  & \multicolumn{1}{l}{TransMIL} & Time(s) \\ \hline
\multirow{4}{*}{TCGA-IDH}   & 10   &0.9&   {$79.3_{\pm0.7}$}  &  $6.3_{(\textcolor{green}{\times1.4})}$   &   {$78.6_{\pm0.5}$}  &   $10.6_{(\textcolor{green}{\times1.14})}$    &   {$75.8_{\pm1.3}$}  &$22.0_{(\textcolor{green}{\times1.07})}$   \\
                       & 20   &1.7&  {$80.4_{\pm0.5}$}   &  $6.5_{(\textcolor{green}{\times1.3})}$     &   {$80.7_{\pm0.5}$}  & $10.8_{(\textcolor{green}{\times1.12})}$     &   {$77.1_{\pm1.6}$}   & $22.1_{(\textcolor{green}{\times1.06})}$      \\
                       & 50   &4.1&   {$82.6_{\pm0.1}$}  &  $8.1_{(\textcolor{green}{\times1.1})}$      &   {$83.3_{\pm0.5}$} & $11.1_{(\textcolor{green}{\times1.09})}$  &   {$79.0_{\pm0.3}$}  &  $23.0_{(\textcolor{green}{\times1.03})}$\\
                       & Full  &4.9& {$84.8_{\pm0.7}$} & 9.0 & {$85.8_{\pm0.1}$} & 12.1 & {$80.0_{\pm0.1}$}& 23.5 \\ \hline
\multirow{4}{*}{Camelyon16} & 10   & 0.32   &$84.1_{\pm0.4}$&$3.1_{(\textcolor{green}{\times1.7})}$&{$84.9_{\pm0.4}$}&$4.7_{(\textcolor{green}{\times1.6})}$& {$81.5_{\pm0.3}$}&$8.4_{(\textcolor{green}{\times1.6})}$\\
                       & 20   & 0.62  &{$84.6_{\pm0.9}$}&$3.5_{(\textcolor{green}{\times1.5})}$&{$85.4_{\pm0.7}$}&$5.1_{(\textcolor{green}{\times1.4})}$&{$82.5_{\pm0.4}$}&$8.6_{(\textcolor{green}{\times1.6})}$\\
                       & 50   & 1.6   &{$86.7_{\pm0.7}$} &$4.2_{(\textcolor{green}{\times1.2})}$& {$87.5_{\pm0.6}$}&$5.9_{(\textcolor{green}{\times1.2})}$&{$83.8_{\pm0.8}$}&$8.9_{(\textcolor{green}{\times1.6})}$\\
                       & Full  & 16.75 & {$88.9_{\pm0.4}$} &5.8&  {$89.8_{\pm0.9}$} &7.3&  {$85.3_{\pm0.3}$} &13.8\\\hline
\multirow{4}{*}{\begin{tabular}[c]{@{}c@{}}UniToPath\\ (DR)\end{tabular}} & 10  &0.22& {$61.6_{\pm0.3}$}    &$1.82_{(\textcolor{green}{\times1.10})}$&  {$62.1_{\pm1.1}$}   &$4.68_{(\textcolor{green}{\times1.04})}$& {$54.2_{\pm1.4}$}       &$5.55_{(\textcolor{green}{\times1.05})}$\\
                       & 20  &0.43& {$64.4_{\pm1.4}$}    &$1.83_{(\textcolor{green}{\times1.10})}$&  {$66.3_{\pm1.9}$}    &$4.74_{(\textcolor{green}{\times1.03})}$&  {$59.1_{\pm1.9}$}     &$5.61_{(\textcolor{green}{\times1.04})}$\\
                       & 50  &1.04& {$67.1_{\pm0.9}$}    &$1.89_{(\textcolor{green}{\times1.06})}$&  {$68.9_{\pm1.9}$}     &$4.77_{(\textcolor{green}{\times1.02})}$& {$61.4_{\pm0.9}$}       &$5.68_{(\textcolor{green}{\times1.02})}$\\
                       & Full &28.66 & {$68.8_{\pm0.9}$} &2.01&  {$70.3_{\pm1.3}$} &4.86&  {$62.9_{\pm0.3}$} &5.82\\\hline
\multirow{4}{*}{\begin{tabular}[c]{@{}c@{}}UniToPath\\ (PT)\end{tabular}} & 10 & 0.71&  {$83.4_{\pm0.7}$} &$2.22_{(\textcolor{green}{\times1.2})}$& {$82.8_{\pm1.2}$}    &$4.74_{(\textcolor{green}{\times1.10})}$&   {$80.5_{\pm1.3}$}   &$6.72_{(\textcolor{green}{\times1.04})}$\\
                       & 20  &0.82&  {$84.8_{\pm0.7}$}   &$2.26_{(\textcolor{green}{\times1.2})}$& {$84.1_{\pm0.1}$}     &$4.76_{(\textcolor{green}{\times1.09})}$&  {$81.9_{\pm0.7}$}    &$6.75_{(\textcolor{green}{\times1.04})}$\\
                       & 50  &2.45& {$87.1_{\pm1.2}$}   &$2.40_{(\textcolor{green}{\times1.1})}$&{$87.5_{\pm2.0}$}     &$4.92_{(\textcolor{green}{\times1.06})}$& {$83.4_{\pm0.7}$}    &$6.82_{(\textcolor{green}{\times1.03})}$\\
                       & Full &34.60& {$90.0_{\pm0.9}$} &2.68& {$91.0_{\pm0.7}$}&5.20&  {$84.7_{\pm1.2}$}  &7.01\\\hline
\end{tabular}
\caption{Slide-level classification results on TCGA-IDH and Camelyon16. We compared different MIL methods.}
\label{table:slide}
\end{table*}

\noindent\textbf{Slide-level classification.}
Though we distilled synthetic patches from the original patch datasets, we can also use them for slide-level tasks where we apply the same algorithm to every single slide to obtain a condensed representation of a slide.
Slide-level classification in Whole Slide Image (WSI) datasets often employs Multiple Instance Learning (MIL).
To conduct MIL training, each slide is treated as a bag and its extracted patches are treated as bag instances.
Here, we leveraged the learned synthetic patches from each slide as bag instances to train several commonly used MIL baselines including CLAM \cite{campanella2019clinical}, DSMIL \cite{li2021dual}, and TransMIL \cite{shao2021transmil}. Subsequently, we evaluated their performance on real datasets.
Specifically, both CLAM and TransMIL used a ResNet50 for feature extraction from the bag instances, followed by either a global pooling operator \cite{campanella2019clinical} or a Transformer \cite{shao2021transmil} for instance feature aggregation. 
In DSMIL, we employed self-supervised learning for enhanced feature extraction and incorporated an attention mechanism for feature aggregation.
The results, as shown in Table \ref{table:slide}, demonstrate the compatibility of our synthetic patches with MIL methods, affirming that our synthetic patches contain informative features from the original dataset that MIL methods can successfully extract and aggregate. 


\noindent\textbf{In-domain cross-centre transfer learning.}
Models that exhibit excellent performance in one setting may demonstrate degraded performance elsewhere due to the differences in population, clinical setting, and user behaviour. This is usually known as the transportability problem in cross-centre studies \cite{coiera2022evidence}. These models are expected to improve through fine-tuning with cross-centre datasets. However, this presents a challenge as centres may be unable to share the raw data due to privacy concerns and governance policies attached to the data. 

Here, we consider a practical scenario in which two medical centres work collaboratively on a given task.
Suppose each centre has collected its own data and would like to share it for collaboration.
Some common issues would occur when using real large-scale datasets, such as the performance of trained models on the new data, storage requirements, and privacy concerns.
To show that our synthetic samples can be an efficient alleviation to these issues, we simulate such in-domain cross-centre transfer learning using the two medical centres (Radboud(C1), Utrecht(C2)) in Camelyon16 dataset.
\begin{table*}[t!]
\scriptsize
\centering
\begin{tabular}{ccc|cccccccc}
\hline
\multicolumn{3}{c|}{\multirow{3}{*}{Source $\Rightarrow$ Target}}  & \multicolumn{8}{c}{Target}   \\ \cline{4-11} 
\multicolumn{3}{c|}{}  & \multicolumn{8}{c}{C2}                                                                     \\ \cline{4-11} 
\multicolumn{3}{c|}{}  & \multicolumn{4}{c|}{Patch-level}                       & \multicolumn{4}{c}{Slide-level}   \\ \hline
\multicolumn{1}{c|}{}                        & \multicolumn{1}{c|}{}                    & Img/Cls & Acc & \#Epochs & Time(s) & \multicolumn{1}{c|}{Storage} & Acc & \#Epochs & Time(s) & Storage \\ \hline
\multicolumn{1}{c|}{\multirow{4}{*}{Source}} & \multicolumn{1}{c|}{\multirow{4}{*}{C1}} & 10      &$94.5_{\pm0.5}$&5&0.23& \multicolumn{1}{c|}{$1e^{-3}$G}&$94.2_{\pm1.2}$&5&1.8&0.20G\\
\multicolumn{1}{c|}{}                        & \multicolumn{1}{c|}{}                    & 20      &$95.7_{\pm0.4}$&5&0.38& \multicolumn{1}{c|}{$2e^{-3}$G}&$94.5_{\pm1.2}$&5&1.9&0.37G\\
\multicolumn{1}{c|}{}                        & \multicolumn{1}{c|}{}                    & 50      &$96.8_{\pm0.2}$&5&1.04& \multicolumn{1}{c|}{$6e^{-3}$G}&$95.8_{\pm0.6}$&5&2.1&0.93G\\
\multicolumn{1}{c|}{}                        & \multicolumn{1}{c|}{}                    & Full    & $97.8_{\pm0.2}$&5&1640& \multicolumn{1}{c|}{5.12G}&$97.0_{\pm0.6}$& 10&2.8&5.12G\\ \hline
\multicolumn{1}{c|}{}                        & \multicolumn{1}{c|}{}                    &         & \multicolumn{8}{c}{C1}                                                                     \\ \hline
\multicolumn{1}{c|}{\multirow{4}{*}{Source}} & \multicolumn{1}{c|}{\multirow{4}{*}{C2}} & 10      &$94.3_{\pm0.1}$&5&0.23& \multicolumn{1}{c|}{$1e^{-3}$G}        &$95.9_{\pm0.1}$&5& 1.5&0.13G\\
\multicolumn{1}{c|}{}                        & \multicolumn{1}{c|}{}                    & 20      &$95.3_{\pm0.5}$&5&0.38& \multicolumn{1}{c|}{$2e^{-3}$G}        &$96.8_{\pm0.6}$&5&1.6&0.25G\\
\multicolumn{1}{c|}{}                        & \multicolumn{1}{c|}{}                    & 50      &$96.4_{\pm0.2}$&5&1.04& \multicolumn{1}{c|}{$6e^{-3}$G}        &$97.5_{\pm0.4}$& 5&1.9&0.64G\\
\multicolumn{1}{c|}{}                        & \multicolumn{1}{c|}{}                    & Full    &$98.1_{\pm0.1}$&5&1596& \multicolumn{1}{c|}{11.63G}        &$97.0_{\pm0.0}$&10& 3.6 &11.63G\\ \hline
\end{tabular}
\caption{Results of in-domain cross-centre transfer learning. Given a model pre-trained on the source dataset, we record its performance and training time for fine-tuning on the target datasets.}
\label{table:centre}
\end{table*}

Suppose the target centre receives the data collected from the source centre (Source $\Rightarrow$ Target) and performs initial investigations by conducting patch-level and slide-level classifications using their own pre-trained models (\emph{e.g.,} ResNet18 for patch-level classification and CLAM for slide-level classification).
The results, as presented in Table \ref{table:centre}, highlight several key advantages of utilising condensed synthetic source domain data over the full source domain data in real clinical practice: 1) Rapid Fine-Tuning: A pre-trained model on the target domain can be swiftly fine-tuned on the newly provided source domain, achieving comparable performance to one fine-tuned on the entire source domain dataset.
2) Enhanced Data Transfer Efficiency: Our synthetic samples significantly reduce storage requirements, leading to more efficient data transfer.

\noindent\textbf{Limitations.}
Histo-DD is task-specific, implying that the distilled samples may not be readily transferable to other tasks, such as image segmentation or detection. 
Here, we present one possible solution. Inspired by the success of continual learning, we could formulate the learning of task-agnostic synthetic samples as a task incremental learning task. 
Specifically, we plan to incrementally introduce a variety of task types, such as classification and segmentation, during the training phase of a set of synthetic samples. 
\section{Conclusion}
This paper introduces a dataset distillation method (Histo-DD) tailored for learning synthetic samples from histopathology datasets.
The proposed method can condense large-scale histopathology datasets consisting of patches into a small set of highly condensed and informative synthetic patches.
During the learning process, we incorporate stain normalisation to further make the algorithm more suitable for histopathology datasets.
We have shown that synthetic patches offer a data-efficient alternative to original large-scale datasets.
Once trained, they enable efficient training for both patch-level and slide-level classifiers. Additionally, they are adaptable to complex tasks like in-domain cross-centre transfer learning.
In future work, our research will focus on refining dataset distillation algorithms, to create task-agnostic synthetic samples.

\clearpage  

%
%
\bibliographystyle{splncs04}
\bibliography{main}
\end{document}


\title{Supplementary Material} 

\titlerunning{Abbreviated paper title}

\author{First Author\inst{1}\orcidlink{0000-1111-2222-3333} \and
Second Author\inst{2,3}\orcidlink{1111-2222-3333-4444} \and
Third Author\inst{3}\orcidlink{2222--3333-4444-5555}}

\authorrunning{F.~Author et al.}

\institute{Princeton University, Princeton NJ 08544, USA \and
Springer Heidelberg, Tiergartenstr.~17, 69121 Heidelberg, Germany
\email{lncs@springer.com}\\
\url{http://www.springer.com/gp/computer-science/lncs} \and
ABC Institute, Rupert-Karls-University Heidelberg, Heidelberg, Germany\\
\email{\{abc,lncs\}@uni-heidelberg.de}}

\maketitle
\section{Dataset Details}
\begin{table}[]
\centering
\begin{tabular}{c|cc|cc}
\hline
                                                                        & \multicolumn{2}{c|}{\# Patches} & \multicolumn{2}{c}{\# Slides} \\ \hline
                                                                        & Train          & Test          & Train         & Test         \\ \hline
TCGA-IDH                                                                    &17,686&2,310 &1,191&149\\ \hline
Camelyon16                                                              &1,066,988&567,510&270&129\\ \hline
\begin{tabular}[c]{@{}c@{}}UniToPath\\ (DG)\end{tabular} &5,631& 2,277&160&70\\ \hline
\begin{tabular}[c]{@{}c@{}}UniToPath\\ (PT)\end{tabular}        &6,889&2,647&208&88\\ \hline
\end{tabular}
\caption{Train/test splits of each dataset. All train/test patches are obtained from the corresponding slides in the train/test sets.}
\label{table:dataset}
\end{table}
The numbers of training/testing patches/slides are provided in Table \ref{table:dataset}. For TCAG-IDH, the patches are cropped at 10$\times$ magnification level, for Camelyon16, the patches are cropped at 5$\times$ magnification level and for UniToPath, the patches are cropped at 20$\times$ magnification level. As the magnification level decreases, the number of cropped patches increases. This variation in magnification levels demonstrates that our Histo-DD is capable of operating across different scales.
\section{Visualisation Results}
\begin{figure}[!t]
\centering
\includegraphics[scale=.35]{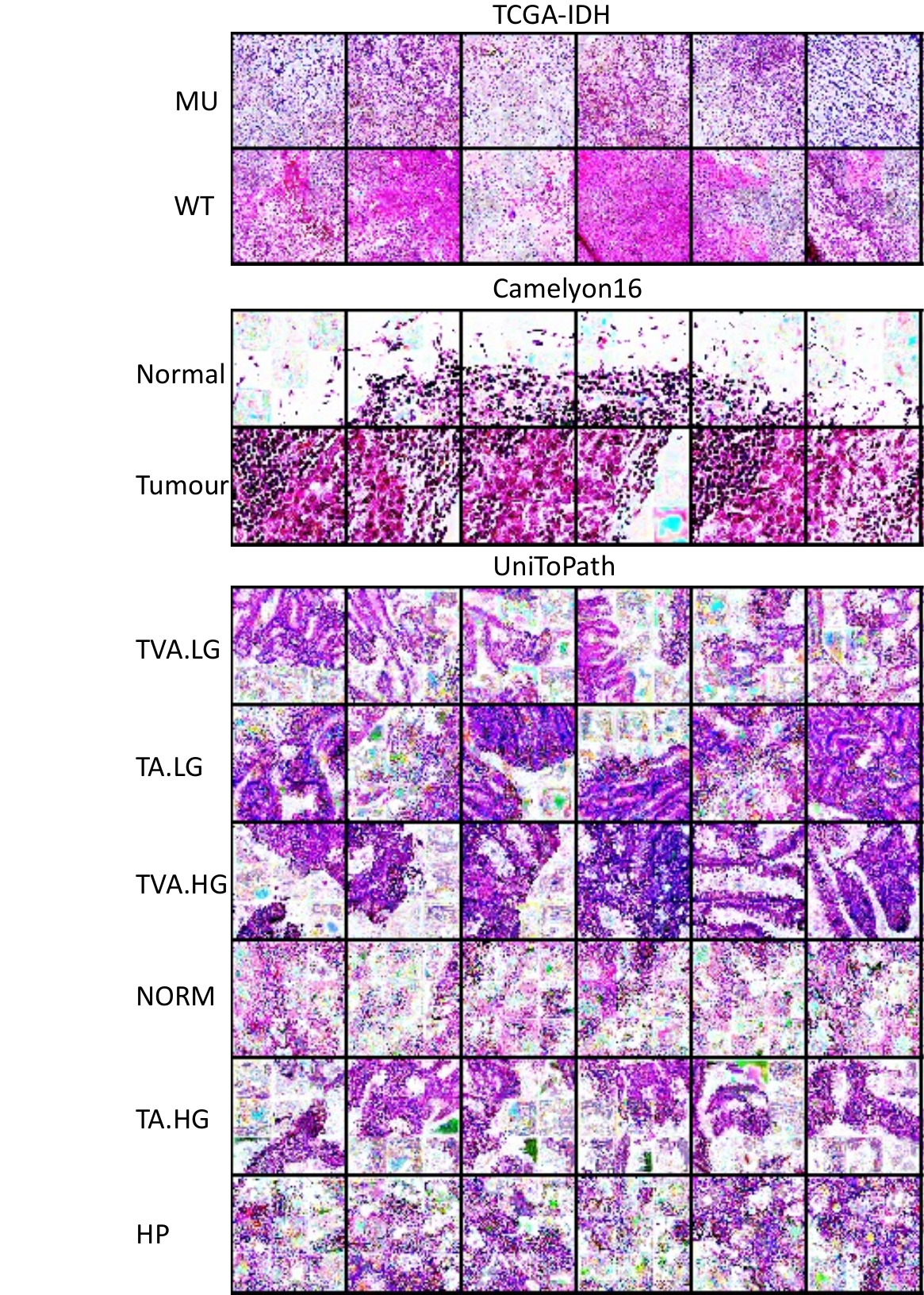}
\caption{Visualisation of condensed image patches.}
\label{fig:vis}
\end{figure}
We provide more visualisation results for 10 patches/class settings in Figure \ref{fig:vis}. 
In cases where the patient identifiers are discernible, they could lead to unintended patient identification. 
In addressing this concern, Histo-DD involves transforming these images into synthetic variants. As shown in Figure \ref{fig:vis}, while these synthetic versions maintain the essential diagnostic characteristics necessary for machine learning applications, they simultaneously obscure or alter specific patient-related details. This process effectively reduces the risk of privacy breaches, as the altered images no longer contain clear identifiers that could be traced back to individual patients. 
\section{Sensitivity study on hyper-parameters.} 
\begin{figure*}[!t]
\centering
\includegraphics[scale=.35]{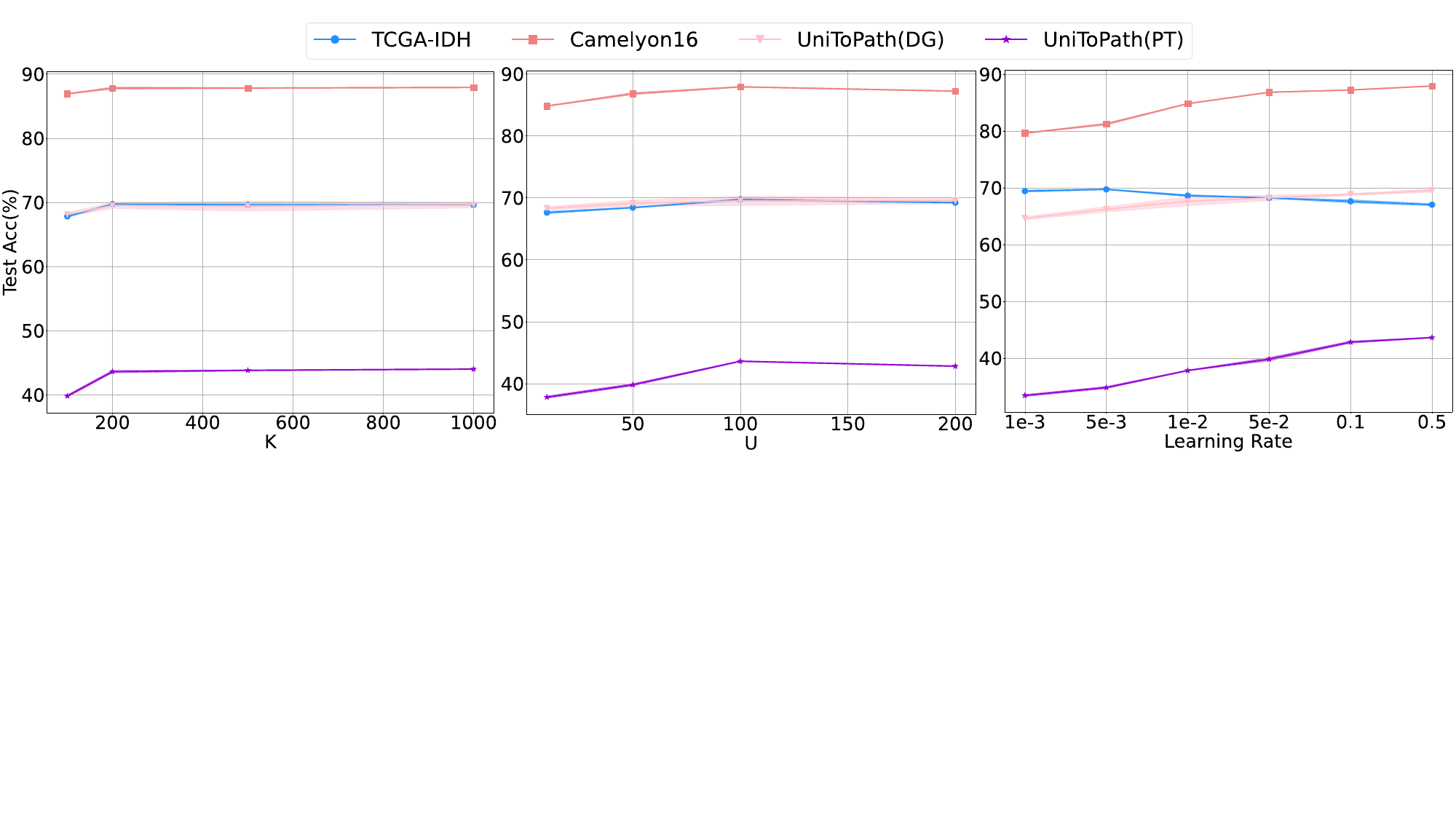}
\caption{Testing accuracy for different numbers of outer loop ($K$), inner loop ($U$), and learning rate when learning 10 images/class.}
\label{fig:hyper}
\end{figure*}
We analysed the sensitivity of different hyper-parameters including the number of outer loops ($K$), the number of inner loops ($U$), and the learning rate. 
The testing performance of a ResNet18 model trained on 10 img/cls is shown in Figure \ref{fig:hyper}. 
Once the number of synthetic samples is determined, we notice that our algorithm is not sensitive to different values for $K$ and $U$. For example, for 10 img/cls, increasing $K$ from 200 to 1000 only provides marginal performance improvement but requires much longer training, thus we set $K=200$ and $U=100$ as the default value for 10 img/cls.
Moreover, the result in Figure \ref{fig:hyper} (c) indicates that the learning rate has a larger impact on the performance. We argue that a smaller learning rate is preferred for simpler datasets (\emph{e.g.,} single-scale, small-scale), and a larger learning rate is preferred when learning from large-scale or multi-scale datasets.

\section{More ablations on the usage of stain normalisation.} 
\begin{table}[]
\centering
\begin{tabular}{cc|c|cccc}
\hline
\multicolumn{2}{c|}{Train} & Test & \multicolumn{4}{c}{Test Accuracy (\%)}                   \\ \hline
SN-Pre   & SN-DSA   & SN   & IDH & Cam16 & UTP (DG) & UTP (PT) \\ \hline
    &   \checkmark  &   \checkmark   &  $\mathbf{69.7_{\pm0.2}}$       &    $\mathbf{86.8_{\pm0.1}}$        &       $\mathbf{69.5_{\pm0.8}}$         &     $\mathbf{43.6_{\pm0.1}}$           \\
    &   \checkmark       &      &          &            &                &                \\ 
 \checkmark   &          &  \checkmark    &  $64.9_{\pm0.2}$       &            &                &                \\
  \checkmark    &          &      &   $67.1_{\pm0.3}$       &            &                &                \\\hline
\end{tabular}
\label{tab:snab}
\caption{Ablation studies on different stain normalisation schemes for 10 synthetic img/cls on different datasets. For training, we either conduct stain normalisation as a pre-processing step on the real image datasets (\textbf{SN-Pre}) or integrate stain normalisation with DSA (\textbf{SN-DSA}). Moreover, we compare results with/without stain normalisation during testing on TCGA-IDH (\textbf{IDH}), Camelyon16 (\textbf{Cam16}) and UniToPath (\textbf{UTP}).}
\end{table}
\section{Extra Use Case: Continual Learning}
\begin{figure}[t!]
\centering
\includegraphics[scale=.24]{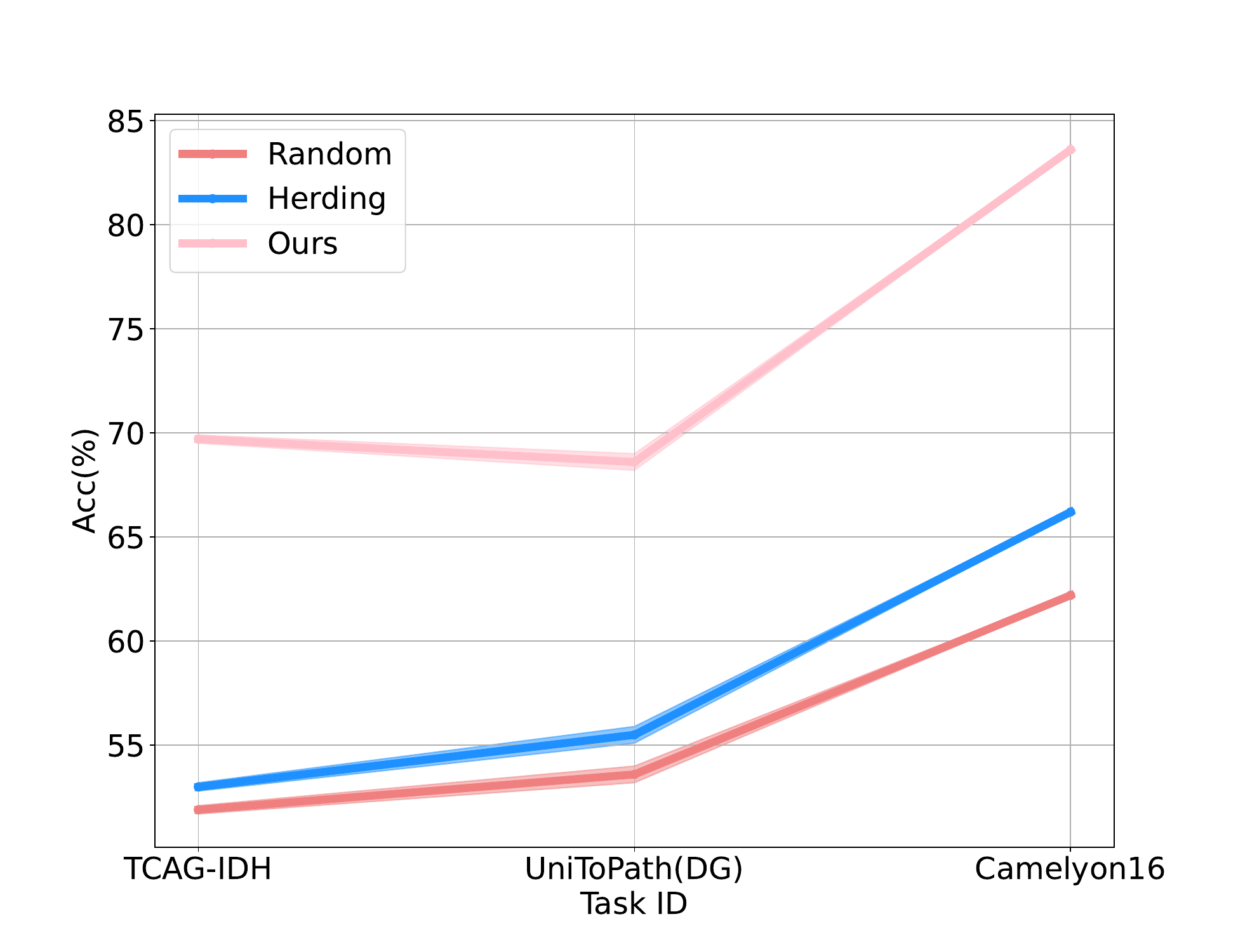}
\caption{Continual learning performance.}
\label{fig:cl}
\end{figure}

We address a more challenging scenario in which tasks from each dataset are incrementally introduced to a model, with the goal of achieving a higher overall accuracy at the conclusion of each stage. In this context, we have arranged the datasets in the following order: TCGA-IDH$\rightarrow$UniToPath(DG)$\rightarrow$Camelyon16, effectively framing it as a task incremental continual learning problem.
Following previous works \cite{zhao2023dataset}, we used GDumb \cite{prabhu2020greedy} as the continual learning baseline which stores training samples in memory and trains the model from scratch using the stored samples when a new task is introduced. 
This setup ensures that the performance of the continual learning method relies solely on the quality of the stored memory.
We compared three memory construction approaches: 1) random selection \cite{prabhu2020greedy}, 2) herding selection \cite{belouadah2020scail,rebuffi2017icarl}, and 3) using synthetic samples produced by Histo-DD. 
Here, we presented the results by selecting 10 samples per class to construct the memory and using ResNet18 for training and testing.
The results, depicted in Figure \ref{fig:cl}, reveal that the synthetic samples generated by Histo-DD (Ours) outperform the other two coreset selection methods, indicating that the synthetic samples generated by Histo-DD are more informative.
%
%
\bibliographystyle{splncs04}
\bibliography{main}